\documentclass[runningheads]{llncs}
\usepackage{dialogue}
\usepackage{tabularx}
\usepackage{amssymb}
\usepackage{amsmath} 
\usepackage{algorithm}
\usepackage{algpseudocode}
\usepackage{float}
\usepackage{mdframed}
\usepackage{xcolor}
\usepackage{threeparttable}
\usepackage[T1]{fontenc}
\usepackage{graphicx}
\usepackage{booktabs}
\usepackage{enumitem}
\usepackage{orcidlink}
\newcommand{\revision}[1]{\textcolor{black}{#1}}

\begin{document}
\title{Understanding Teacher Revisions of Large Language Model-Generated Feedback}
\titlerunning{Teacher Revisions of LLM-Generated Feedback}
\author{Conrad Borchers\inst{1}\orcidlink{0000-0003-3437-8979} \and
Luiz Rodrigues\inst{2}\orcidlink{0000-0003-0343-3701} \and
Newarney Torrezão da Costa\inst{3}\orcidlink{0000-0002-4954-176X} \and
Cleon Xavier\inst{3}\orcidlink{0000-0002-7617-5283} \and
Rafael Ferreira Mello\inst{4}\orcidlink{0000-0003-3548-9670}}
\authorrunning{Borchers et al.}

\institute{Carnegie Mellon University\\
\email{cborcher@cs.cmu.edu}\\
\and
Federal Technological University of Paraná -- UTFPR\\
\email{luizrodrigues@utfpr.edu.br}\\
\and
Instituto Federal Goiano -- IF Goiano\\
\email{\{newarney,cleon.junior\}@ifgoiano.edu.br}
\and
Federal Rural University of Pernambuco -- UFRPE\\
\email{rafael.mello@ufrpe.br}
}

\maketitle 
\begin{abstract}
Large language models (LLMs) increasingly generate formative feedback for students, yet little is known about how teachers revise this feedback before it reaches learners. Teachers' revisions shape what students receive, making revision practices central to evaluating AI classroom tools. We analyze a dataset of 1,349 instances of AI-generated feedback and corresponding teacher-edited explanations from 117 teachers. We examine (i) textual characteristics associated with teacher revisions, (ii) whether revision decisions can be predicted from the AI feedback text, and (iii) how revisions change the pedagogical type of feedback delivered. First, we find that teachers accept AI feedback without modification in about 80\% of cases, while edited feedback tends to be significantly longer and subsequently shortened by teachers. Editing behavior varies substantially across teachers: about 50\% never edit AI feedback, and only about 10\% edit more than two-thirds of feedback instances. Second, machine learning models trained only on the AI feedback text as input features, using sentence embeddings, achieve fair performance in identifying which feedback will be revised ($AUC$=0.75). Third, qualitative coding shows that when revisions occur, teachers often simplify AI-generated feedback, shifting it away from high-information explanations toward more concise, corrective forms. Together, these findings characterize how teachers engage with AI-generated feedback in practice and highlight opportunities to design feedback systems that better align with teacher priorities while reducing unnecessary editing effort.
\end{abstract}

\section{Introduction}

Large language models (LLMs) have been increasingly used to support assessment and feedback in education, particularly in writing and tutoring \cite{xiao2025,thomas2025tutors}. Their appeal is largely pragmatic: LLMs can generate timely, detailed feedback at scale, potentially reducing teacher workload and expanding access to formative support \cite{xavier2025human}. As these systems transition from experimental prototypes to routine classroom tools, however, a critical question is not only whether models can generate plausible feedback, but how educators actually use, interpret, and revise that feedback before it reaches students.

This question is especially salient given growing evidence that LLM-generated feedback can be pedagogically misaligned. Prior work shows that LLMs may produce feedback that varies systematically with superficial cues such as gendered language \cite{du2025benchmarking}, or that favors overly direct, overscaffolded explanations that conflict with learning sciences principles \cite{borchers2025can,stamper2024enhancing}, potentially undermining learning outcomes \cite{fan2024beware}. In classroom settings, these shortcomings are not merely theoretical: teachers act as gatekeepers who decide whether, how, and to what extent AI-generated feedback is delivered to students.

Teacher mediation matters because educators may mitigate some of the adverse effects of LLM-generated feedback described above \cite{borchers2025can,du2025benchmarking}, such as misalignment with effective instructional principles \cite{borchers2025can}. However, it cannot be assumed that such mediation will consistently improve feedback quality. Prior research shows that teachers, like other practitioners, may hold beliefs about learning that are not supported by empirical evidence, such as the persistent myth of learning styles \cite{zambrano2025prevalence}. At the same time, emerging evidence suggests that teachers often forward AI-generated feedback with little or no modification \cite{xavier2025empowering}, raising the stakes of understanding when and why revisions occur. If teachers rarely edit AI-generated feedback, its pedagogical impact is largely determined by the model's defaults, the prompt, and the system in which the AI is embedded. If teachers do not frequently edit, then ineffective or biased model behavior (which can be hard to detect) \cite{du2025benchmarking} in AI feedback can affect students. Evidence suggests that LLMs often misalign with effective instruction \cite{borchers2025can}.

Most evaluations of LLM-based feedback instead judge output quality in isolation, such as by comparing responses to rubrics or evidence-based principles \cite{borchers2025can,thomas2025tutors}, while ignoring the human-in-the-loop processes that shape what students ultimately see. As a result, it remains unclear how AI-generated feedback functions once teachers decide whether and how to pass it on, and where current systems may impose avoidable cognitive or editorial burden. We address this gap by empirically examining how teachers revise AI-generated formative feedback through large-scale analysis of revision behavior, semantic similarity measures, predictive modeling, and theory-informed coding of feedback types. In doing so, this work contributes to AIED by identifying opportunities to design feedback systems that better align with teacher priorities while reducing unnecessary editing effort.

\section{Related Work}

\subsection{AI Feedback}

AI-supported automated feedback, particularly using LLMs, has been widely studied across educational contexts. Prior work has examined both learner reception and comparative effectiveness relative to human feedback. Letteri et al. \cite{letteri2025enhancing} analyze students’ perceptions of an AI feedback system in data science education, showing that although learners find the feedback helpful, excessive detail can increase cognitive load and divert attention from the task. Nygren et al. \cite{nygren2025ai} empirically compare AI-generated and expert human feedback in a mixed-reality teacher education simulation, finding that AI provides timely support for simpler tasks but lacks the contextual sensitivity required to identify pedagogical opportunities such as missed teachable moments or adaptive lesson structuring.

Other studies emphasize hybrid approaches and the social dimensions of feedback. Khojasteh et al. \cite{khojasteh2025comparing} compare teacher, AI, and combined feedback for second-language writing, demonstrating that hybrid feedback yields greater gains in coherence and grammatical accuracy while also reducing student anxiety and increasing confidence. Nazaretsky et al. \cite{nazaretsky2026gives} focus on feedback attribution rather than content, showing in a large within-subject study that students systematically perceive AI-provided feedback as less credible and lower in quality than equivalent human feedback when its source is disclosed.

Together, these studies establish the instructional promise of AI-generated feedback and the advantages of combining AI with human expertise. Still, they predominantly evaluate feedback as received by students, using performance outcomes or subjective perceptions. In contrast, the present study shifts attention to teachers’ engagement with AI-generated feedback itself. We examine how teachers revise LLM-produced feedback, which textual features predict revision decisions, and how pedagogical feedback types change through this process, thereby identifying systematic misalignments between current AI-generated feedback and teachers’ instructional priorities.

\subsection{Teacher Engagement with AI Feedback}

Prior work consistently shows that teachers do not treat AI feedback as authoritative; instead, they exercise judgment in interpreting and adapting it. For instance, pre-service teachers reflect on the adequacy and contextual sensitivity of AI feedback as they make pedagogical decisions \cite{nygren2025ai}. Related work positions teachers not only as evaluators but also as designers of AI feedback systems. One study found that instructors configure how AI-generated suggestions are presented to students in data science education, shaping both the timing and framing of feedback \cite{letteri2025enhancing}. This highlights that teachers actively orchestrate how automated feedback reaches students.

Similarly, past work examines preservice English teachers who compare their own feedback on authentic student texts with GenAI-generated suggestions \cite{fredriksson2025gift}. Participants report concerns about the excessive length and density of AI feedback, prompting selective adoption based on accuracy, relevance, and pedagogical tone. Teachers frequently revise or discard suggestions to prevent cognitive overload and preserve motivational balance.

Together, these studies depict teachers as critical mediators who adapt AI feedback to pedagogical goals. The present work extends this literature by offering an automated, quantitative account of such engagement. Analyzing 1,349 instances of AI feedback edited by 117 teachers, we identify systematic revision patterns, including substantial shortening, alongside marked individual variability. These results provide concrete design insights for AI feedback tools that reduce unnecessary effort while strengthening teacher agency.

\subsection{Limitations of AI Feedback and Instruction}

Recent AIED work moves beyond studies of teacher perceptions to examine technical and pedagogical limitations of LLM-generated feedback using natural language processing methods. Borchers and Shou \cite{borchers2025can} show that LLMs struggle to replicate the adaptivity of tutoring systems, exhibiting weak sensitivity to learner errors and student help-seeking. As a result, feedback tends to be overly direct, conflicting with learning science principles such as using open-ended prompts to probe understanding. Relatedly, Du et al. \cite{du2025benchmarking} quantify gender bias in LLM essay feedback through controlled prompt experiments. These findings point to the need for teacher intervention to correct pedagogical incoherence, reduce prompt-driven variability, and realign feedback with effective pedagogy.

Stamper et al. \cite{stamper2024enhancing} similarly identify structural weaknesses in LLM feedback in tutoring contexts, including limited empirical validation of learning outcomes and reliance on fixed-response patterns with minimal adaptation, especially in ill-defined domains. These limitations manifest in feedback that overlooks common student errors, prematurely reveals answers, and fails to account for prior knowledge, scaffolding needs, or multimodal support. Our study builds on this work by providing an integrated empirical account of how teachers systematically reshape the pedagogical character of AI-generated feedback. We identify concrete patterns of correction and refinement that can inform the design of interfaces and LLM behaviors that better reflect instructional intent.

\subsection{The Present Study}

Recent work in AIED argues for evaluation methods that capture subtle but meaningful differences in model behavior \cite{DBLP:conf/aied/KarumbaiahGBA24}. Recent work shows that embedding-based analyses can quantify how nuanced changes in LLM instruction can lead to significant semantic shifts in LLM response distributions \cite{borchers2025can,du2025benchmarking}. We adopt this representational perspective to study teacher revision behavior by embedding AI-generated feedback and teacher-edited versions in a shared semantic space. This allows us to measure alignment, characterize systematic deviations, and analyze variation across educators at scale.

From this framing, we examine how teachers revise AI-generated formative feedback in practice. We ask: \textbf{RQ1}, what textual characteristics are associated with teachers revising AI-generated feedback; \textbf{RQ2}, to what extent can revision decisions be predicted from the AI feedback text alone; and \textbf{RQ3}, when revisions occur, how does the pedagogical type of feedback change. These questions are central because revisions reveal where teachers perceive AI feedback as misaligned with instructional intent. Identifying these patterns can guide the design of feedback systems that better support teacher judgment.

\section{Methods}

\subsection{Dataset and Preprocessing}

\begin{figure}[htpb]
    \centering
    \includegraphics[width=\linewidth]{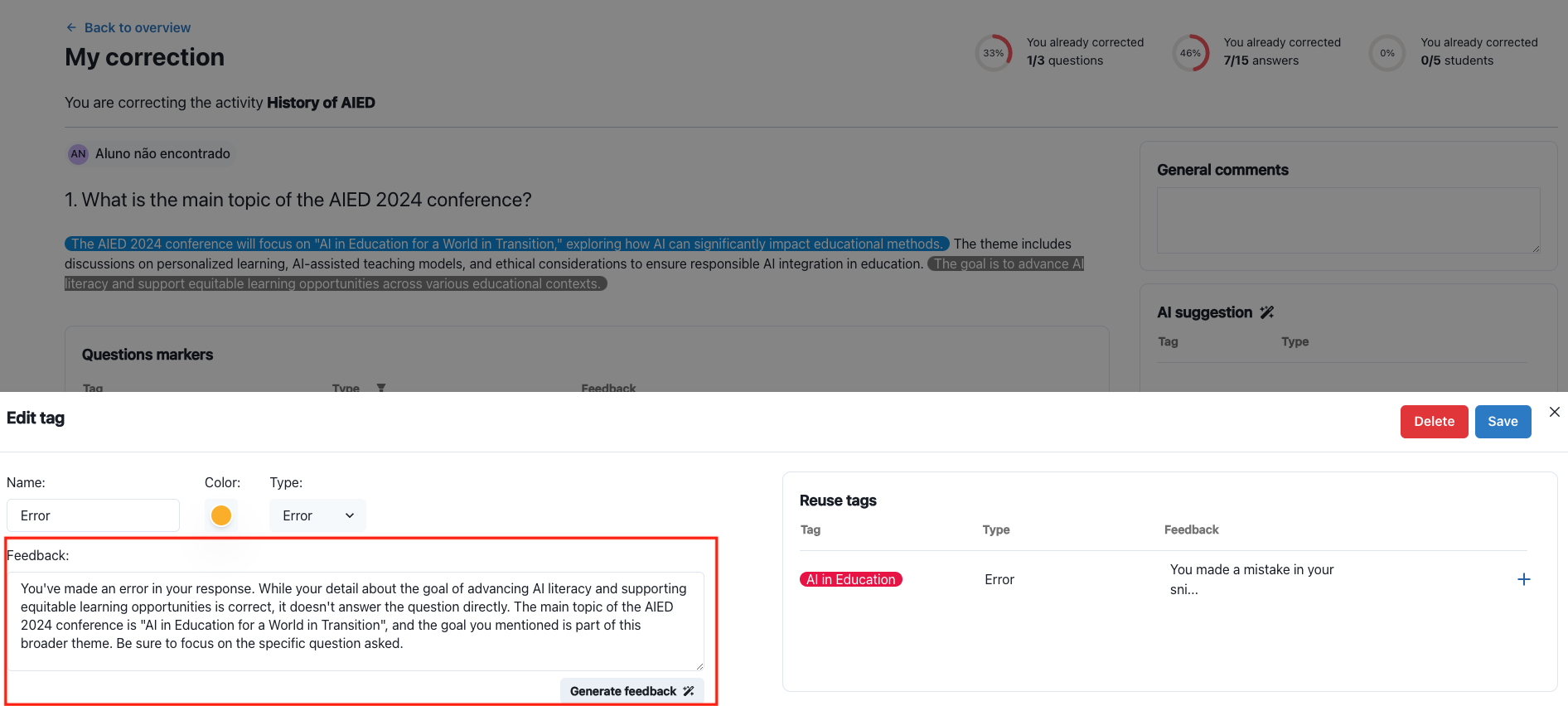}
    \caption{LLM-supported feedback generation. Screenshot of the teacher's tag, evaluation, and feedback module. The red area shows the feedback generated after clicking the Generate Feedback button. Additionally, the screen displays the student's response and two tags created by a teacher.}
    \label{fig:feedback}
\end{figure}

The study draws on log data from Tutoria, an AI-supported feedback platform integrated with Moodle and Google Classroom \cite{Pontual2023tutoria}. The platform supports grading of short, open-ended responses through a co-creation workflow: for each student answer, an LLM generates an initial feedback draft that the teacher can accept as-is, edit, or fully rewrite before delivering it to the student. Teachers also tag responses as correct, partially correct, or incorrect (either for selected spans or the full response); these tags are used as inputs to feedback generation (Fig.~\ref{fig:feedback}). Our dataset pools interaction logs from prior controlled studies of the platform and additional in-the-wild classroom deployments. In total, the dataset includes 1,349 AI-generated feedback drafts and corresponding final teacher-delivered feedback messages authored by 117 teachers. Teacher demographics were available for a subset of the sample: in one set of contributing studies, the participating teachers were 68 Brazilian educators (50 male, 18 female; $M_{age}=37$, $SD=6$). The students who received feedback from teachers similarly span multiple educational contexts and institutions. Contributing contexts included (i) Brazilian high school chemistry classes (periodic table) completed in Moodle, (ii) a controlled experiment in which higher-education instructors provided feedback on open-ended responses in introductory computer science, and (iii) additional extracurricular contexts where computer science students answered questions on topics such as algorithm analysis and computer architecture. Across settings, tasks consisted of comparable short open-ended responses, which teachers evaluated using the platform’s tagging interface and then delivered feedback based on the AI draft. All teachers and students provided informed consent for the research use of their data.

The platform used two large language models during its development. Early versions relied on GPT-4 (gpt-4-0613), while later stages adopted DeepSeek-V3, including multi-agent configurations in the final phase \revision{(for more details, see \cite{neto2025llmagentgrader})}. \revision{Since our analysis} focuses on teachers’ interactions with AI-generated feedback rather than on model quality\revision{, we do not analyze each LLM separately.} Prompts followed established best practices \cite{borchers2025can}, combining explicit instructions, structured outputs, and examples. Each prompt included the question, the student’s response, the teacher-assigned correctness label, and a request for feedback that appropriately praised the response and/or offered guidance toward a more accurate answer. All textual inputs were preprocessed by replacing missing values with empty strings, standardizing format, removing newline characters, and truncating text to a fixed maximum length. These steps were applied uniformly to AI-generated feedback and teacher-authored explanations.

Text was then converted to dense, 384-dimensional vector embeddings using the pre-trained sentence embedding model \texttt{all-MiniLM-L6-v2} from the Sentence Transformers library, which is trained on large-scale natural language inference and paraphrase data \cite{reimers-2019-sentence-bert}. Similar embeddings have been successfully used in past AIED research to understand LLM feedback and labeling instruction \cite{borchers2025can,rodrigues2023question}, successfully ensuring that semantically similar texts have similar vectors.

\subsection{Textual Characteristics Explaining Teacher Revisions (RQ1)}

To characterize textual properties associated with teacher revision behavior, we analyzed surface-level and semantic features of AI-generated feedback. Specifically, we examined feedback length (word count) and semantic similarity between AI feedback and the final teacher explanation, measured using cosine similarity between their embedding representations, following past work \cite{borchers2025can,rodrigues2023question}. These measures capture both the amount of information provided and the extent to which teachers preserved or altered the underlying content of the feedback.

Analyses were primarily descriptive. We compared these characteristics for feedback that was accepted unchanged versus edited. To account for teacher-level differences in editing tendencies, we additionally conducted within-teacher comparisons, using paired nonparametric tests where applicable. These analyses were intended to identify patterns associated with revision behavior rather than to establish causal relationships.

\subsection{Predictive Accuracy for Revision Decisions (RQ2)}

We modeled teacher revision as a binary classification task, predicting whether AI-generated feedback was modified before delivery. To prevent leakage, we constructed a holdout test set comprising one third of the data using a group-based split, with no teacher appearing in both sets \cite{Baker2025BigDataEducation}. Model selection and tuning were conducted on the remaining data via group-based five-fold cross-validation. We evaluated regularized logistic regression, shallow neural networks, and gradient-boosted decision trees, testing 5, 48, and 81 configurations, respectively; full details are available online \cite{teacherrevisions}. Model classes represent increasing complexity, from linear baselines to nonlinear models. Hyperparameters were selected using AUC. \revision{To mitigate outcome class imbalance, logistic regression used class-balanced loss weights, and evaluation emphasized imbalance-robust metrics. Accordingly}, we report AUC, balanced accuracy, and Cohen’s $\kappa$, all standard in AIED \cite{Baker2025BigDataEducation}. Final performance was computed on the holdout set, with 95\% confidence intervals estimated via 10{,}000 group-bootstrap resamples.

\subsection{Feedback Type Differences After Revision (RQ3)}

To examine how teacher revisions changed the pedagogical nature of feedback, we coded AI-generated feedback and final teacher explanations using a theory-informed codebook grounded in prior research on feedback effectiveness \cite{hattie2007power,castro2023understanding}. Consistent with established frameworks \cite{wisniewski2020power}, the codebook distinguished four mutually exclusive categories: Reinforcement/Punishment, comprising evaluative statements with minimal informational content; Corrective feedback, which signals correctness or provides the correct answer with little explanation; High-information feedback, which extends corrective feedback with explanations, strategies, or prompts supporting self-regulation; and Other.

Following the original codebook \cite{wisniewski2020power}, coders assigned a single predominant feedback type to each unique message. \revision{To ease coding workload, we only coded feedback messages with exact duplicates once,} yielding a deduplicated sample of N=834. When AI feedback was accepted without revision, the same code was applied to both the AI output and the final teacher explanation. Feedback coded as \textit{Other} required a brief written justification. Coding was conducted independently, with an initial validation phase after 10\% of the data (n=83) to surface and resolve questions before proceeding. Four coders then independently coded randomly assigned subsets of messages. This procedure aligns with interpretivist qualitative research, where coding reflects theoretically grounded judgment rather than an objective ground truth \cite{blandford2016qualitative}. Consistency was supported by a previously validated codebook with explicit definitions, rules, and examples, and by the validation phase, which ensured shared interpretation before full-scale analysis \cite{blandford2016qualitative}. We open-source our coding instructions for reproducibility \cite{teacherrevisions}.

We analyzed feedback type distributions to examine the relationship between revision status and explanatory form, using an $\alpha=0.05$ significance threshold. To meet test assumptions, low-frequency categories ($<$10\%) were excluded or collapsed (see Section \ref{sec:result_rq3}). For unrevised feedback, a goodness-of-fit test against a uniform distribution assessed whether feedback types occurred equally often, indicating which types were more likely to be revised. For revised feedback, we compared AI-generated and teacher-adjusted distributions using a chi-square test of independence to assess pre-to-post changes in feedback types. Effect sizes (Cohen’s w, Cramér’s V) were reported, and statistically significant shifts were interpreted through qualitative analysis of teacher revisions.

\section{Results}

\subsection{Textual Characteristics Explaining Teacher Revisions (RQ1)}

Descriptive statistics for overall and edited-only cases are summarized in Table~\ref{tab:descriptives}. Across all instances ($N = 1349$), the average cosine similarity between AI feedback and the final teacher-provided explanation was almost perfect ($M = 0.97$, $SD = 0.09$), indicating substantial overlap in content. On average, AI feedback contained 43.2 words, while teacher explanations contained 41.6 words. Teachers accepted AI feedback without modification in 77.8\% of cases ($1050/1349$). Edited feedback was modestly longer than unedited feedback (48.3 vs. 41.9 words), with a mean within-teacher difference of 6.4 words; this difference was statistically significant based on a paired Wilcoxon signed-rank test ($W$ = 460.5, $p = .024$). 

\begin{table}[htpb]
\centering
\caption{Descriptive statistics for AI feedback and final teacher explanations. Values are reported as mean (SD).}
\label{tab:descriptives}
\begin{tabular}{llll}
\toprule
\textbf{Subset} &
\textbf{Similarity} &
\textbf{AI Feedback Length} &
\textbf{Final Feedback Length} \\
\midrule
All feedback &
0.97 (0.09) &
43.24 (19.60) &
41.60 (19.81) \\
Edited feedback only &
0.88 (0.17) &
48.38 (18.09) &
41.00 (19.97) \\
\bottomrule
\end{tabular}
\end{table}

\begin{figure}[htb]
  \centering
  \includegraphics[width=\linewidth]{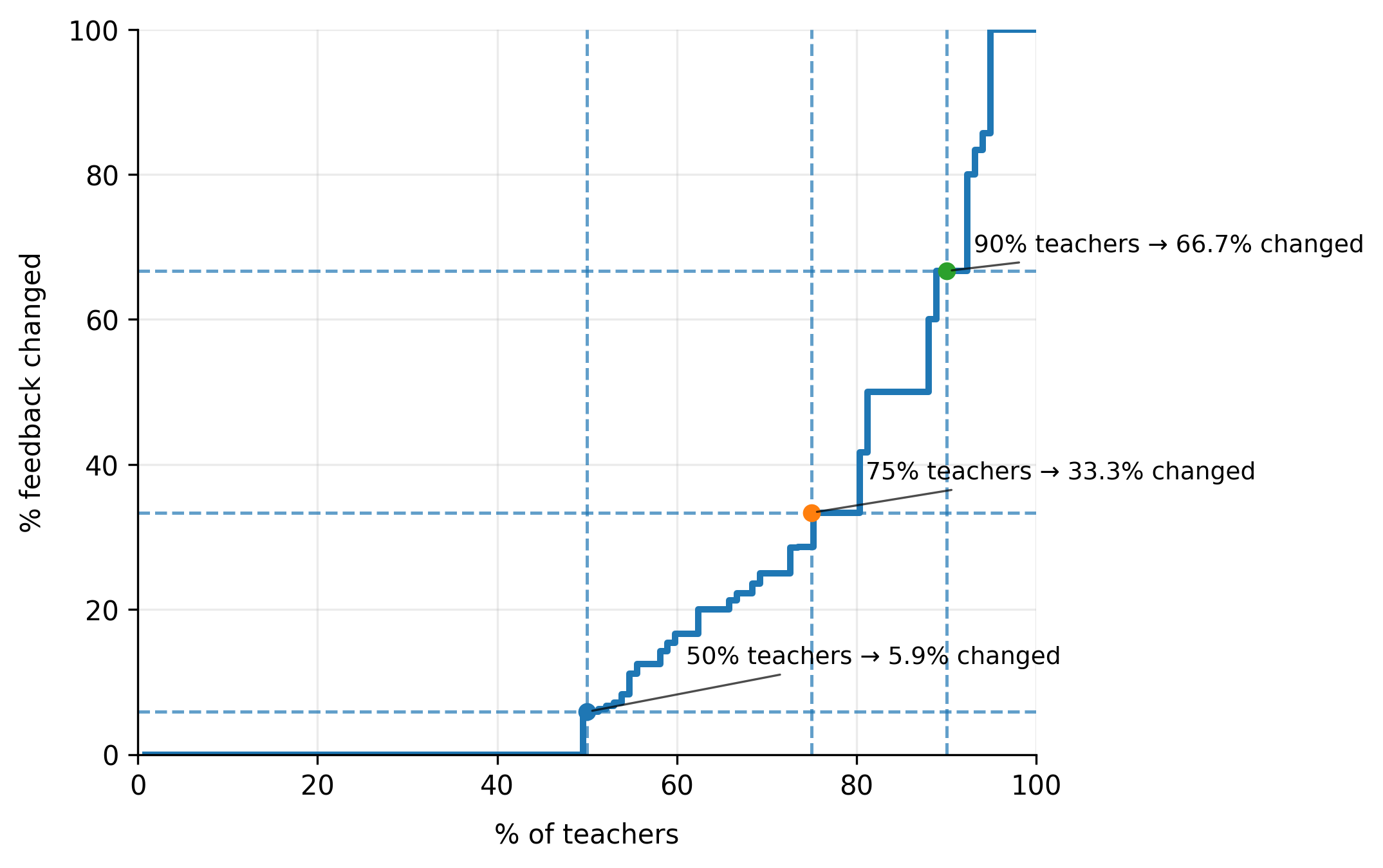}
  \caption{Cumulative distribution of the percentage of messages changed by teachers. The x-axis shows the percentage of teachers (sorted by change rate), and the y-axis shows the percentage of messages modified. Dashed lines indicate teacher percentiles.}
  \label{fig:text_change_cdf}
\end{figure}

We then analyzed editing behavior at the teacher level. Of 117 teachers, 60 (51.3\%) edited AI-generated feedback at least once. Teachers reviewed an average of 11.5 feedback messages ($SD$ = 39.6). Figure~\ref{fig:text_change_cdf} plots the cumulative distribution of editing rates by teacher. Half of teachers edited fewer than 6\% of feedback instances, whereas the upper quartile edited more than 33\%, and the top decile revised roughly two-thirds. This pattern reveals pronounced heterogeneity, with most teachers making minimal changes.

\subsection{Predictive Accuracy for Revision Decisions (RQ2)}

Table~\ref{tab:rq1_models} reports predictive performance on the holdout test set. Across models, performance was modest but consistently above chance (i.e., an $AUC$ of .5). Logistic regression achieved a holdout $AUC$ of .70 [.53, .72], the shallow neural network .74 [.52, .78], and gradient-boosted trees .75 [.54, .77]. Balanced accuracy followed a similar pattern, and Cohen's $\kappa$ indicated close-to-modest performance in most models. Confidence intervals overlapped substantially across models, indicating no reliable performance differences in this dataset.

\begin{table}[htpb]
\centering
\caption{Holdout predictive performance for revision decision models. Values are reported as point estimate [95\% group-bootstrap CI], computed with 10,000 resamples.}
\label{tab:rq1_models}
\resizebox{\textwidth}{!}{%
\begin{tabular}{lcccc}
\toprule
\textbf{Model} &
\textbf{ROC AUC} &
\textbf{Balanced Accuracy} &
\textbf{Accuracy} &
\textbf{Cohen's $\kappa$} \\
\midrule
Logistic Regression &
0.698 [0.534, 0.715] &
0.672 [0.510, 0.688] &
0.733 [0.712, 0.748] &
0.322 [0.015, 0.358] \\
MLP (1 layer) &
0.738 [0.518, 0.781] &
0.535 [0.515, 0.550] &
0.757 [0.740, 0.844] &
0.095 [0.036, 0.134] \\
Gradient-Boosted Trees &
0.746 [0.539, 0.768] &
0.651 [0.487, 0.668] &
0.824 [0.807, 0.864] &
0.387 [-0.039, 0.423] \\
\bottomrule
\end{tabular}%
}
\end{table}

\subsection{Feedback Type Differences After Revision (RQ3)}
\label{sec:result_rq3}

When AI-generated feedback was not revised, feedback was most commonly Corrective ($n$ = 381, 58.1\%), then High-information ($n$ = 275, 41.9\%). A single instance of Reinforcement/Punishment was excluded from further analysis due to its rarity. A chi-square goodness-of-fit test against a uniform distribution indicated that this imbalance was statistically significant, $\chi^2$(1) = 17.13, $p$ $<$ .001, with a small-to-moderate effect size (Cohen's $w$ = 0.16). This pattern suggests that when AI feedback is deemed acceptable by teachers without revision, it is more often corrective than high-information.

When AI-generated feedback was revised, the initial message was significantly more often High-information ($n$ = 106, 59.9\%) than Corrective ($n$ = 71, 40.1\%), $\chi^2$(1) = 6.92, $p$ = .009, with a small-to-moderate effect ($w$ = 0.20). After revision, the distribution of final feedback explanations became more balanced, with 48.0\% High-information and 43.5\% Corrective, alongside a small proportion of Reinforcement/Punishment ($n$ = 13, 7.3\%) and Other feedback (n = 2, 1.1\%). Given the small proportion of Reinforcement/Punishment and Other types, a direct comparison of High-information versus all other feedback types between the initial and final stages revealed a significant shift, $\chi^2$(1) = 4.55, $p$ = .033, with a small association (Cramér's $V$ = 0.12). Human revision reduced the relative prevalence of High-information feedback in favor of other explanatory forms.

To better understand revision patterns, one researcher conducted an exploratory study of feedback messages in which teachers reduced the complexity of AI-generated feedback. In some cases, teachers seemed to disagree with the AI's assessment. For instance, \textit{Congratulations! You were correct in stating that ...} (i.e., high-information) was revised to \textit{That's a very generic and incomplete answer.} (i.e., corrective). Another case was when AI generated high-information feedback in English, which the teacher then revised into a corrective one in Brazilian Portuguese. Additionally, there were cases in which corrective feedback (e.g., \textit{Your answer is correct. You correctly mentioned "devices," which are present in the context of the question referring to input and output devices. Keep it up!}) was replaced by a Reinforcement/Punishment one due to plagiarism concerns (i.e., \textit{Your answer is correct. But be careful of possible plagiarism.}). Lastly, there were cases in which the teacher seemed to disagree with the pedagogical approach of the AI-generated feedback, replacing a high-information one with \textit{Congratulations on correctly identifying that the CPU ...}. Together, these insights suggest that teachers simplified feedback when it conflicted with evaluative judgment, contextual awareness, or feedback efficiency.

\section{Discussion}

This study investigated how teachers revise and deliver LLM-generated formative feedback. By comparing AI feedback with teacher-edited versions, we shift attention from feedback quality in isolation to the human mediation that determines what students ultimately receive. This approach directly answers calls in AIED to examine AI systems in use \cite{DBLP:conf/aied/KarumbaiahGBA24,stamper2024enhancing,borchers2025can}. We synthesize our findings around three contributions: patterns of teacher revision, what these revisions reveal about pedagogical alignment, and implications for feedback systems.

\subsection{Teacher Revision as Selective Mediation}

Teachers accepted AI feedback without modification in nearly 80\% of cases, and more than half of the teachers never edited AI feedback at all. This pattern echoes recent work suggesting that educators often forward AI-generated feedback with minimal intervention \cite{xavier2025empowering}. From an AIED perspective, this result highlights that expecting teachers to continuously and carefully monitor all AI-mediated instructional processes might be unrealistic, particularly in time-constrained classroom settings. Instead, our findings align with the theoretical perspective of human--AI hybrid adaptivity in AIED \cite{holstein2020conceptual}, whereby responsibility for moment-to-moment instructional decisions is distributed between human and system rather than centrally managed by the teacher. In this view, teachers selectively intervene when AI behavior violates pedagogical expectations or contextual constraints.

Notably, revision behavior was highly heterogeneous. Only a few teachers edited a large proportion of feedback messages, in some cases revising more than two-thirds of AI outputs. This variability suggests that teacher mediation is an emergent outcome shaped by individual beliefs, instructional norms, and time constraints. In future work, we recommend further investigating which teacher and contextual attributes explain revision behavior, given substantial variation in teacher trust and self-efficacy for AI \cite{viberg2025explains}.

\subsection{Length, Density, and Pedagogy Trigger Revision}

When teachers revised AI-generated feedback, their revisions followed systematic, interpretable patterns. Addressing RQ1, teacher edits typically reduced length while largely preserving semantic content, suggesting that revisions often functioned as compression rather than correction. This finding aligns with prior qualitative evidence that teachers are concerned about the excessive volume of AI-generated feedback \cite{fredriksson2025gift}. Extending this work, our results provide quantitative empirical evidence that verbosity itself is a trigger for teacher intervention.

Even after teachers reduced the length, the semantic similarity between AI-generated feedback and edited feedback remained high, indicating that most revisions were rhetorical or structural. Qualitative cases nevertheless reveal salient exceptions in which teachers intervened because of disagreement with the pedagogical framing of the feedback (RQ3). Unedited AI feedback was more often high-information than corrective. In contrast, revised feedback was disproportionately high-information and shifted toward a more balanced distribution after editing, suggesting that teachers typically reduced overly explanatory feedback.

This pattern corroborates long-standing concerns regarding the assistance dilemma and the risks of overscaffolding \cite{stamper2024enhancing,koedinger2007exploring}. Prior analyses of LLM-generated tutoring and feedback show that models frequently default to explicit explanations and direct answers, even when indirect prompts or opportunities for productive struggle may better support learning \cite{borchers2025can}. Our findings suggest that teachers sometimes counteract this tendency by simplifying feedback, reasserting evaluative judgment, or narrowing the instructional focus to what they deem pedagogically essential. Reducing explanation can lower cognitive load, but it may also strip away conceptual scaffolds necessary for learner sensemaking \cite{koedinger2007exploring}. Teachers’ revisions reveal how this trade-off is actively negotiated, exposing divergences between AI defaults and human pedagogical judgment in practice.

Future research should examine how teachers calibrate LLMs’ tendency to overscaffold, which can offload learner sensemaking and impair learning \cite{fan2024beware}, or instead overcorrect and provide insufficient support \cite{koedinger2007exploring}. Classroom experiments comparing edited and unedited AI feedback, combined with learning analytics that quantify scaffolding from language, offer a promising path forward \cite{borchers2025disentangling}.

\subsection{Predictability of Revision and the Limits of Text-Only Signals}

Addressing RQ2, our predictive modeling results show that it is possible to identify, at above-chance levels, which AI feedback messages will be revised using only the AI-generated text as input. At the same time, model performance remained modest (about 0.70-0.75 $AUC$ and close to 0.4 $\kappa$), with wide confidence intervals (though the lower bounds generally exceeded the at-chance level of 0.5 $AUC$) and no clear advantage for more complex architectures. Substantively, the modest performance of text-only models shows that revision decisions are not determined solely by the properties of AI-generated feedback. Prior work suggests that teachers’ feedback practices are shaped by additional factors such as the student’s response and error type, the teacher’s grading intent and evaluative stance \cite{borchers2025disentangling}, disciplinary conventions, and situational constraints such as time pressure. None of these signals were available to the models in the present study. Incorporating such contextual information—potentially through multimodal or task-aware models—may substantially improve predictive performance \cite{wong2025rethinking}. More accurate models, in turn, could support practical applications, such as prioritizing feedback instances likely to require human revision, thereby reducing unnecessary teacher effort while preserving pedagogical control.

\subsection{Implications for AI Feedback Systems}

Our findings carry clear implications for AIED. The prevalence of unedited feedback indicates that default AI behavior is instructionally consequential; evaluation paradigms that assume routine teacher correction therefore underestimate the effects of bias, verbosity, and pedagogical misalignment \cite{du2025benchmarking,fan2024beware,borchers2025can}. When teachers do revise, they primarily shorten or reframe rather than rewrite. At the same time, marked variation across teachers underscores the limits of uniform feedback generation. Systems that adapt to patterns in a teacher’s prior edits may better align with classroom practice without constraining professional judgment.

These findings call for evaluating AI feedback as a co-created artifact shaped through teacher interaction, rather than focusing solely on raw model output \cite{thomas2025tutors}. Learning analytics \revision{and future research} can capture revision traces and interaction costs and relate them to student outcomes, providing a foundation for designing feedback systems that better support teachers in practice \cite{borchers2025disentangling}.

\subsection{Limitations and Future Directions}

This study examines a retrospective dataset of short, open-ended tasks within a single platform, and revision behavior may differ across assessment types, interface designs, and educational contexts. Treating revision as a binary outcome further masks meaningful variation between minor edits and substantive rewrites. \revision{Further,} acceptance of feedback should not be equated with agreement or trust, as it may instead reflect time pressure\revision{, pedagogical intent toward particular students,} or editing costs. \revision{It is also possible that embeddings derived from LLMs fine-tuned on educational data would provide representations with greater predictive validity. Finally, controlled experiments that hold the LLM and prompts constant would enable clearer attribution of revision differences to specific models, an analysis beyond the scope of the present study.}

\section{Conclusion}

This study argues that evaluating LLM-generated formative feedback in isolation is no longer sufficient for AIED. In practice, feedback reaches students through a mediated pipeline in which teachers are the final authors. Analysis of AI-generated feedback reviewed by 117 teachers shows that this mediation is selective: most feedback is delivered unchanged, making default model behavior instructionally consequential. When revisions occur, however, they are systematic and revealing, signaling where LLMs misalign with teachers’ priorities.

We conceptualize teacher revision as a measurable indicator of pedagogical alignment and show that it can be studied at scale using surface features and representational similarity. Teachers most often compress and simplify AI drafts, shifting from explanatory to concise corrective feedback. This pattern captures a core tension in feedback design and highlights teacher judgment as an essential object of inquiry for AIED. Moving forward, our field should pay more attention to teacher revisions as a core object of inquiry for AIED and a practical lever for designing systems that better reflect teacher preferences.


\bibliographystyle{splncs04}
\bibliography{main} 

\end{document}